\documentclass[10pt,twocolumn,letterpaper]{article}

\usepackage{cvpr}
\usepackage{times}
\usepackage{epsfig}
\usepackage{graphicx}
\usepackage{amsmath}
\usepackage{amssymb}
\usepackage{pgfplots}
\usepackage{setspace}
\usepackage[boxed,linesnumbered]{algorithm2e} 
\pgfplotsset{compat=newest}

\usepackage{calrsfs}
\DeclareMathAlphabet{\pazocal}{OMS}{zplm}{m}{n}
\newcommand{\argmax}{\mathop{\mathrm{argmax}}\limits}   

\newcommand{\Ib}{\pazocal{I}}

\usepackage[pagebackref=true,breaklinks=true,letterpaper=true,colorlinks,bookmarks=false]{hyperref}

 \cvprfinalcopy 

\def\cvprPaperID{0303} 

\begin{document}

\title{Weakly Supervised Cascaded Convolutional Networks}

\author{
    {  Ali Diba$^{1}$,  Vivek Sharma$^{1}$, Ali Pazandeh$^{1,2}$, Hamed Pirsiavash$^{3}$ and Luc Van Gool$^{1,4}$}\\
    {\normalsize {$^{1}$ESAT-PSI, KU Leuven, $^{2}$Sharif Tech., $^{3}$UMBC, $^{4}$CVL, ETH Z\"{u}rich}} \\ 
     \tt\small \{firstname.lastname\}@esat.kuleuven.be, hpirsiav@umbc.edu
 }


\maketitle

\begin{abstract}
Object detection is a challenging task in visual understanding domain, and even more so if the supervision is to be weak. Recently, few efforts to handle the task without expensive human annotations is established by promising deep neural network. A new architecture of cascaded networks is proposed to learn a convolutional neural network (CNN) under such conditions. We introduce two such architectures, with either two cascade stages or three which are trained in an end-to-end pipeline. The first stage of both architectures extracts best candidate of class specific region proposals by training a fully convolutional network. In the case of the three stage architecture, the middle stage provides object segmentation, using the output of the activation maps of first stage. The final stage of both architectures is a part of a convolutional neural network that performs multiple instance learning on proposals extracted in the previous stage(s). Our experiments on the PASCAL VOC 2007, 2010, 2012 and large scale object datasets, ILSVRC 2013, 2014 datasets show improvements in the areas of weakly-supervised object detection, classification and localization.  

\end{abstract}

\section{Introduction}
The ability to train a system that detects objects in cluttered scenes by only naming the objects in the training images, without specifying their number or their bounding boxes, is understood to be of major importance. Then it becomes possible to annotate very large datasets or to automatically collect them from the web. 

Most current methods to train object detection systems assume strong supervision \cite{fastRCNN, fasterRCNN, SSD}. Providing both the bounding boxes and their labels as annotations for each object, still renders such methods more powerful than their weakly supervised counterparts. Although the availability of larger sets of training data is advantageous for the training of convolutional neural networks (CNNs), weak supervision as a means of producing those has only been embraced to a limited degree. 

The proposed weak supervision methods have come in some different flavours. One of the most common approaches \cite{cinbis} consists of the following steps. The first step
generates object proposals. The second stage extracts features from the proposals. And the final stage applies multiple instance learning (MIL) to the features and finds the box labels from the weak bag (image) labels. This approach can thus be improved by enhancing any of its setps. For instance, it would be advantageous if the first stage were to produce more reliable - and therefore fewer - object proposals.   

\begin{figure}[t]
\centering
{\includegraphics[width=220pt]{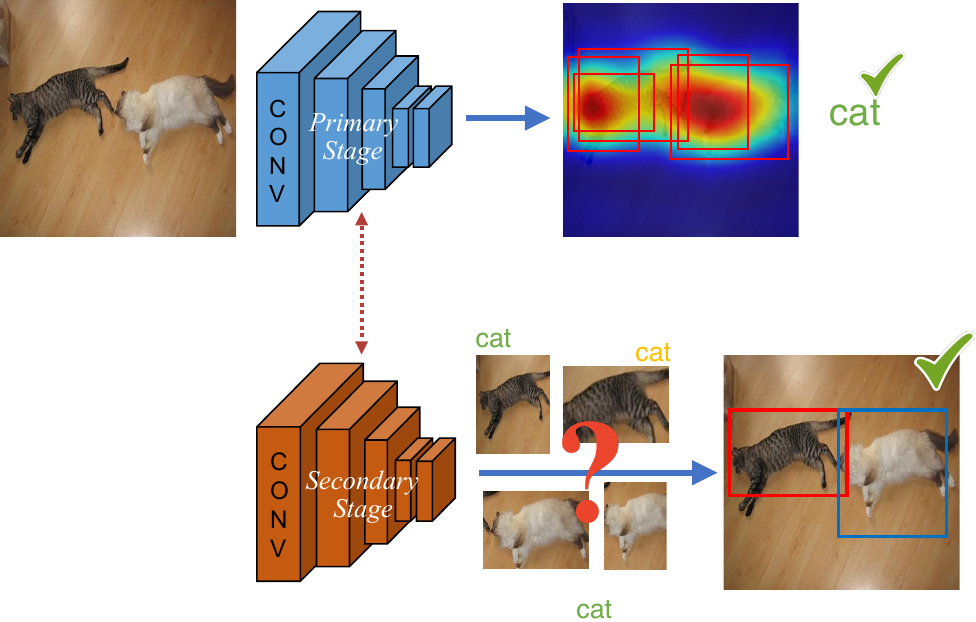} 
} 
\caption{\textbf{Weakly Supervised Cascaded Deep CNN:} Overview of the proposed cascaded weakly supervised object detection and classification method. Our cascaded networks take images and existing object labels to find the best location of objects samples in each of images. Trained networks based on these location is capable of detecting and classifying objects in images, under weakly supervision circumstances. } 
\label{fig:gradient}
\end{figure}

It is the aforementioned approach that our weak supervision algorithm also follows. To improve the detection performance, object proposal generation, feature extraction, and MIL are trained in a cascaded manner, in an end-to-end way.  
%
%
We propose two architectures. The first is a two stage network. The first stage extracts class specific object proposals using a fully convolutional network followed by a global average (max) pooling layer. The second stage extracts features from the object proposals by a ROI pooling layer and performs MIL. Given the importance of getting better object proposals we added a middle stage to the previous architecture in our three stage network. This middle stage performs a class specific segmentation using the input images and the extracted objectness of the first stage. This results in more reliable object proposals and a better detection.

The proposed architecture improves both initial object proposal extraction and final object detection. In the forward sense, less noisy proposals indeed lead to improved object detection, due to the non-convexity of the cost function. In the reverse, backward sense, due the weight sharing between the first layers of both stages, training the MIL on the extracted proposals will improve the performance of feature extraction in the first convolutional layers and as a result will produce more reliable proposals.

Next, we review related works in section 2 and discuss our proposed method in section 3. In section 4 we explain the details of our experiments, incl. the dataset and complete set of experiments and results.

\section{Related works}

\paragraph{Weakly supervised detection:}
In the last decade, several weakly supervised object detection methods have been studied using multiple instance learning algorithms~\cite{bilen14,bilen15,siva,song14}. To do so they define images as the bag of regions, wherein they assume the image labeled positive contains at least one object instance of a certain category and  an image labeled negative do not contain an object from the category of interest. The most common way of weakly supervised learning methods often work by selecting the candidate positive object instances in the positive bags, and then learning a model of the object appearance using appearance model. Due to the training phase of the MIL problem alternating between out of bag object extraction and training classifiers, the solutions are non-convex and as a result is sensitive to the initialization. In practice, a bad initialization is prone to getting the solution stuck in a local optima, instead of global optima. To alleviate this shortcoming, several methods try to improve the initialization~\cite{song14a,does10, Siva12,siva} as the solution strongly depends on the initialization, while some others focus on regularizing the optimization strategies ~\cite{bilen14,bilen15,cinbis}. Kumar et al.~\cite{kumar} employ an iterative self-learning strategy to employ harder samples to a small set of initial samples at training stage. Joulin et al.~\cite{joulin} use a convex  relaxation of soft-max loss in order to minimize the prone to get stuck in the local minima. Deselaers et al.~\cite{does10}  initialize the  object locations via the objectness score. Cinbis et al.~\cite{cinbis} split the training date in a multi-fold manner for escaping from getting trapped into the local minima. In order to have more robustness from poor initialization, Song et al.~\cite{song14}  apply Nesterov's smooting technique to latent SVM formulation~\cite{dpm}. In~\cite{song14a}, the same authors initialize the object locations based on sub-modular clustering method. Bilen et al.~\cite{bilen14} formulates the MIL to softly label the object instances by regularizing the latent object locations based on penalizing unlikely configurations. Further in~\cite{bilen15}, the authors extend their work~\cite{bilen14} by enforcing similarity between object windows via regularization  technique.  Wang et al.~\cite{wang14}  employ probabilistic latent semantic analysis on the windows of positive samples to select the most discriminative clusters that represents the object category.  As a matter of fact, majority of the previous works~\cite{reed14,sukh14} use a large collection of noisy object proposals to train their object detector. In contrast, our method only focuses on a very few clean collection of object proposals that are far more reliable, robust, computationally efficient, and gives better performance.

\paragraph{Object proposal generation:}
 In~\cite{Nguyen,Pandey}, Nguyen et al. and Pandey et al. extract dense regions of candidate proposals from an image using an initial bounding box. To handle the problem of not being able to generate enough candidate proposals because of fixed shape and size, object saliency~\cite{does10,Siva12,siva} based approaches were proposed to extract region proposals. Following this, generic objectness measure~\cite{alex10} was employed to extract region proposals. Selective search algorithm~\cite{SS}, a segmentation based object proposal generation was proposed, which is currently among the most promising techniques used for proposal generation. Recently, Ghodrati et al.~\cite{diba} proposed an inverse cascade method using various CNN feature maps to localize object proposals in a coarse to fine manner.

\paragraph{CNN based weakly supervised object detection:}
In view of the promising results of CNNs for visual recognition, some recent efforts in weakly supervised classification have been based on CNNs. Oquab et al.~\cite{Oquab14} improved feature discrimination based on a pre-trained CNN. In~\cite{laptev15}, the same authors improved the performance further by incorporating both localization and classification on a new CNN architecture. Bilen et al.~\cite{bilen14} proposed a CNN-based convex optimization method to solve the problem to escape from getting stuck in local minima. Their soft similarity between possible regions and clusters was helpful in improving the optimization. Li et al.~\cite{li16} introduced a class-specific object proposal generation based on the mask out strategy of \cite{bazzani}, in order to have  a reliable initialization. They also proposed their two-stage algorithm, classification adaptation and detection adaptation.  


\begin{figure*}[ht]
 \centering
 \includegraphics[width=500pt]{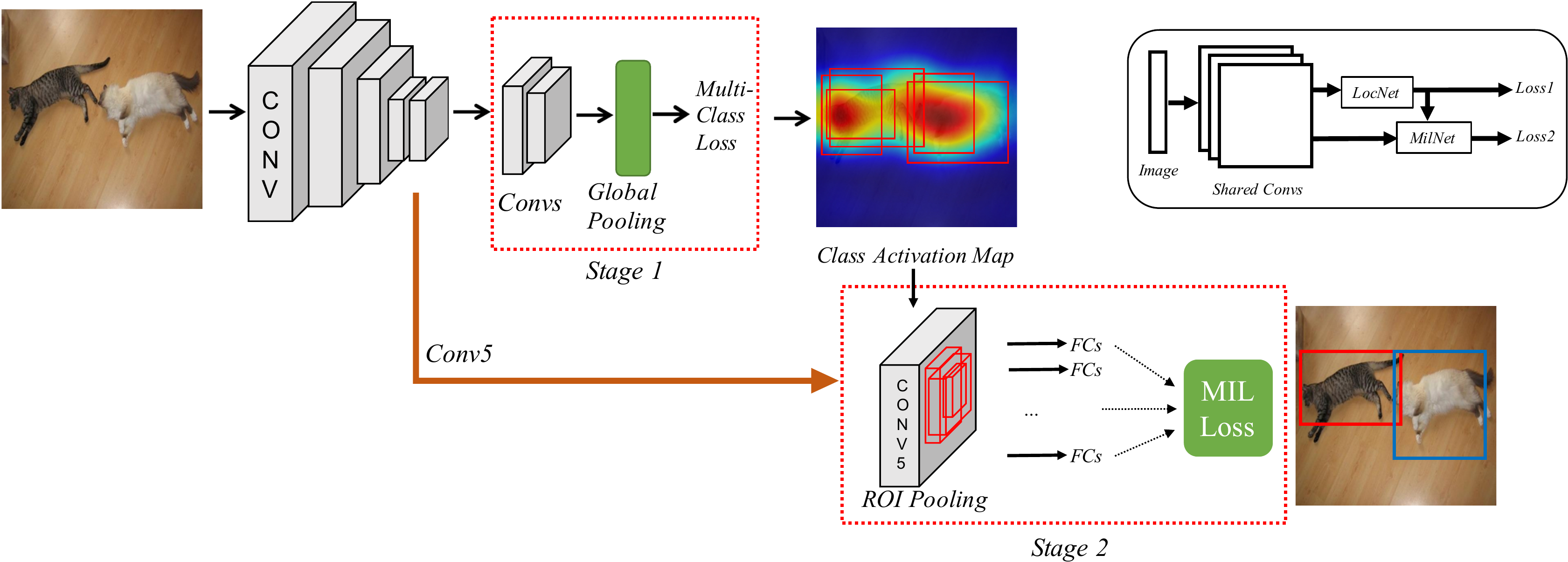}
 \caption{\textbf{WCCN (2stage):} The pipeline of end-to-end 2-stage cascaded CNN for weakly supervised object detection. Inputs to the network are images, labels and unsupervised object proposals. First stage learns to create a class activation map based on object categories to make some candidate boxes for each instance of objects. Second stage picks the best bounding box among the candidates to represent the specific category by multiple instance learning loss.}
  \label{fig:2}
 \end{figure*} 

\section{Proposed Method}

This section introduces our weak cascaded convolutional networks (WCCN) for object detection and classification with weak supervision. Our networks are designed to learn multiple different but related tasks all together jointly. The tasks are classification, localization, and multiple instance learning. We show that learning these tasks jointly in an end-to-end fashion results in better object detection and localization. The goal is to learn good appearance models from images with multiple objects where the only manual supervision signal is image-level labels. Our main contribution is improving multiple object detection with such weak annotation. To this end, we propose two different cascaded network architectures. The first one is a 2-stage cascade network that first localizes the objects and then learns to detect them in a multiple instance learning framework. Our second architecture is a 3-stage cascade network where the new stage performs semantic segmentation with pseudo ground truth in a weakly supervised setting.

\subsection{Two-stage Cascade}
As mentioned earlier, there are only a few end-to-end frameworks with deep CNNs for weakly supervised object detection. In particular, there is not much prior art on object localization without localization information in the supervision. 
Suppose we have a dataset $\it{\Ib}$ with $\it{C}$ classes in \textit{N} training images. The set is given as $\Ib=\{(I^1,y^1),..., (I^N,y^N) \}$ where $\it{I}$ are images and $\mathbf{y} = [y_{1},..., y_{C}] \in \{0,1\}^C$ are vectors of labels indicating the presence or absence of each class in a given image.

In the proposed cascaded network, the initial fully-convolutional stage learns to infer object location maps based on the object labels in the given images. This stage produces some candidate boxes of objects as input to the next stage. The second stage selects the best boxes through an end-to-end multiple instance learning in the network. 

\textbf{First stage (Location network):} The first stage of our cascaded model is a fully-convolutional CNN with a global average pooling (GAP) or global maximum pooling (GMP) layer, inspired by \cite{gap_paper}. The training yields the object location or `class activation' maps, that provide candidate bounding boxes. 
In order to learn multiple classes and to address the issue of multiple categories label for single image~\cite{laptev15}, we use an independent loss function for each class in this branch of the CNN architecture, so the loss function is the sum of \textit{C} binary logistic regression loss functions.

 \begin{figure*}[ht]
 \centering
 \includegraphics[width=500pt]{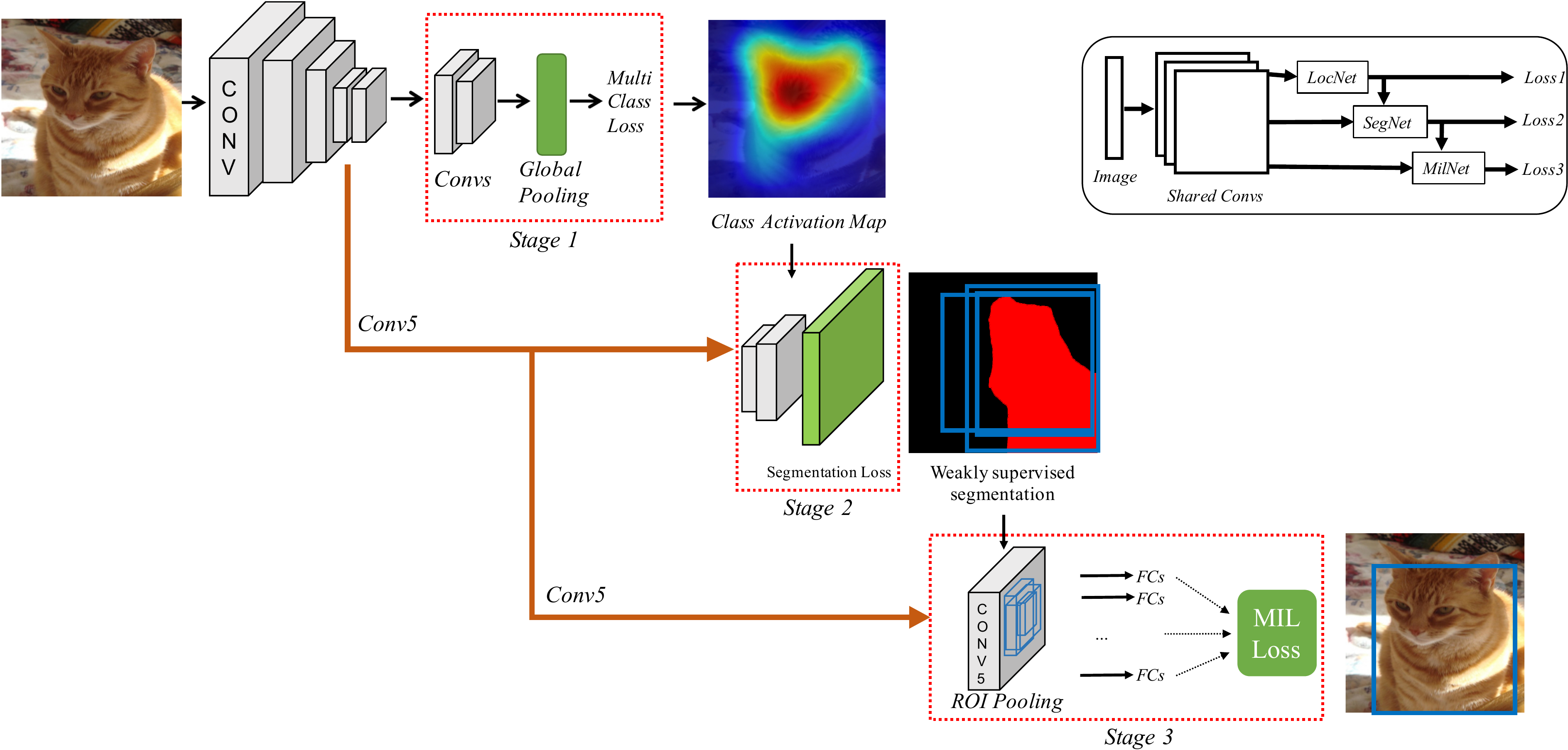}
 \caption{\textbf{WCCN (3stage):} The pipeline of end-to-end 3-stage cascaded CNN for weakly supervised object detection. For this cascaded network, we designed new architecture to have weakly supervised segmentation as second stage, so first and third stages are identical to the stages of the previous cascade. The new stage will improve the selecting candidate bounding boxes by providing more accurate object regions.}
  \label{fig:3}
 \end{figure*}
 
\textbf{Second stage (MIL network):} The goal of the second stage is to select the best candidate boxes for each class from the outputs of the first stage, using MIL. To obtain an end-to-end framework, we incorporate an MIL loss function. For multiple instance learning, we consider $x_{c}=\{x_{j}|j = 1,2,..., n\}$ as a bag for instances of image \textit{I} and each of \textit{x} is one of the candidate boxes and label sets of $y_{x}=\{y_{i}|y_{i} \in \{0,1\}, i =1,...,C\}$ for the bag where ${\sum_{i=1}^C y_{i} = 1}$ and the reason is that each positive bag should belong to one specific object category. Using bounding boxes of instances, we extract CNN representation for each box by ROI-pooling layer \cite{fastRCNN}: $f = \{f_{ij}\} \in \Re^{C\times n}$. So we define and probabilities and loss as: 
\begin{equation}
\begin{split}
Score(I,f)_{i}=\max(f_{i1},..., f_{in}) \\
P(I,f_{i}) = \frac{exp(Score(I,f_{i})_{i})}{\sum_{{k=1}}^C exp(Score(I,f_{k})_{k})} \\ 
L_{MIL}(P,y) = -\sum_{{i=1}}^C y_{i} log(P(I,f_{i}))
\end{split}
\label{eq:1}
\end{equation}

The weights for \textit{conv1} till \textit{conv5} are shared between the two stages. For the second stage, we have additional two fully connected layers and a score layer for learning MIL task.

\textbf{End-to-End Training:} The whole cascade with two loss functions is learned jointly by end-to-end stochastic gradient descent optimization. The total loss function of the cascaded network is:
\begin{equation}
\begin{split}
L_{Total} = L_{GAP}(Labels(W))+ \qquad  \qquad  \qquad  \qquad\\
L_{MIL}(Labels(W)|candidate Boxes(W)).
\label{eq:2}
\end{split}
\end{equation}
\noindent where $\it{W}$ contains all network parameters. We set the hyperparameter balancing two loss functions to $1$. We suspect cross-validation on this hyperparameter can improve the results in the experiments.

\subsection{Three-stage Cascade}
In this section, we extend our 2-stage cascaded model by another stage that adds object segmentation as another task. We believe more information about the objects' boundary learned in a segmentation task can lead to acquisition of a better appearance model and then better object localization. For this purpose, our new stage uses another form of weak supervision to learn a segmentation model, embedded in the cascaded network and trained along with other stages. This extra stage will help the multi-loss CNN to have better initial locations for choosing candidate bounding boxes to pass to the next stage. So this new cascade has three stages: \textbf{first stage}, similar to previous cascade is a CNN with global pooling layer; \textbf{second stage}, fully convolutional network with segmentation loss; \textbf{third stage}, multiple instance learning with corresponding loss.

\textbf{New stage (Segmentation Loss):} Inspired by \cite{pointwise, weakSeg}, we propose to use a weakly supervised segmentation network which uses an object point of location and also label as supervisory signals. Incorporation of initial location of object from previous stage (location network) in the segmentation stage can obtain more meaningful object location map. The weak segmentation network uses the results of the first stage as supervision signal (i.e., pseudo ground truth) and learns jointly with the MIL stage to further improve the object localization results.

To calculate the loss for this stage, we define $s_{ic}$ for the CNN score for pixel $\it{i}$ and class $\it{c}$ in image $\it{I}$. Eq.\ref{eq:3}, shows the softmax for class $\it{c}$ at pixel $\it{i}$.

\begin{equation}
S_{ic}={exp(s_{ic})}/{\sum_{{k=1}}^C exp(s_{ik})}
\label{eq:3}
\end{equation}

Considering $\it{y}$ as the label set for image $\it{I}$, the loss function for the weakly supervised segmentation network is given by:
\begin{equation}
\begin{split}
L_{Seg}(S,G,y)=-\sum_{{i=1}}^C y_{i} log(S_{t_{c}c})-{\sum_{i \in I_{s}}\alpha_{i} log(S_{t_{c}G_{i}})}\\
By\quad t_{c}=\argmax_{i \in I}{S_{ic}}\qquad \qquad \quad \qquad
\end{split}
\label{eq:4}
\end{equation}
\noindent where the first term is used for image-level label supervision and second term is for the set of labeled pixels in $\it{I_{s}}$. ${G_{i}}$ is the supervision map for the segmentation which is obtained from first stage of cascade and not annotated by human. $\alpha_{i}$ denotes the score of importance for each pixel at the map which is calculated in the last stage.

Output of this stage is a set of candidate bounding boxes of objects for pushing to next stage of the CNN cascade which uses multiple instance learning to choose the most accurate box as the representative of object category.  In the experiments, we show that learning this extra task as another stage of cascade can improve performance of the whole network as a weakly supervised classifier.

\textbf{End-to-End Training:}
Similar to the last cascade, the total loss in Eq.\ref{eq:5} is calculated by simply adding all three loss terms. We learn all parameters of the network jointly in an end-to-end fashion.
\begin{equation}
\begin{split}
L_{Total} = L_{GAP}(Labels(W))+ \qquad  \qquad  \qquad \qquad \\
L_{Seg}(Map(W)|Point(W))+ \qquad \qquad \quad \\
L_{MIL}(Labels(W)|candidate Boxes(W)).
\label{eq:5}
\end{split}
\end{equation}

\subsection{Object Detection Training}
Our cascaded network can be used in an object detector pipeline in two ways. The direct way is to use the network after training as the main part of detection. The network is capable of targeting the location and label of the existing object instances in the image. So we can push images and unsupervised object proposals to the cascade and operate all the stages for labeling, localizing and finding the best boxes for each of the object category or rejecting boxes as non-object. 

Second way is to use best extracted location of objects in the training phase as new ground-truths (GT) and train an efficient supervised object detector pipeline like R-CNN or Fast-RCNN \cite{fastRCNN}. So these obtained bounding boxes are acting as pseudo GT and replace the manual annotations. In both cases, at the testing time, we extract object proposals with EdgeBoxes \cite{edgebox} and use the train networks in either case to detect objects among the pool of proposals. Non-max-suppression is also used to clarify final decisions on the boxes and throwing away redundant cases. In the experiments, we show good results for both these methods.   

\section{Experiments}
In the following section, we discuss full details of our methods and experiments which we applied on object detection and classification in weakly supervised manner. We introduce datasets and also analyze performance of our approaches on them in many aspects of evaluation.

\begin{table*}[!htbp]
  \centering
  \resizebox{\textwidth}{!}{
  \begin{tabular}{  |c | c c c c c c c c c c c c c c c c c c c c| c| } 
 \hline
 Method & aero & bike & bird & boat & bottle & bus & car & cat & chair & cow & table & dog & horse & mbike & person & plant & sheep & sofa & train & tv & mAP\\ [0.5ex]
 \hline
 
Bilen et al. \cite{bilen14} & 42.2 &43.9 &23.1 &9.2 &12.5 &44.9 &45.1 &24.9 &8.3 &24.0 &13.9 &18.6 &31.6 &43.6 &7.6 &20.9 &26.6 &20.6 &35.9 &29.6 &26.4\\ [0.5ex]

Bilen et al. \cite{bilen15} & 46.2 &46.9 &24.1 &16.4 &12.2 &42.2 &47.1 &35.2 &7.8 &28.3 &12.7 &21.5 &30.1 &42.4 &7.8 &20.0 &26.8 &20.8 &35.8 &29.6 &27.7\\ [0.5ex]

Cinbis et al. \cite{cinbis} & 39.3 &43.0 &28.8 &20.4 &8.0 &45.5 &47.9 &22.1 &8.4 &33.5 &23.6 &29.2 &38.5 &47.9 &20.3 &20.0 &35.8 &30.8 &41.0 &20.1 &30.2\\[0.5ex]

Wang et al. \cite{wang14} &48.8 &41.0 &23.6 &12.1 &11.1 &42.7 &40.9 &35.5 &11.1 &36.6 &18.4 &35.3 &34.8 &51.3 &17.2 &17.4 &26.8 &32.8 &35.1 &45.6 &30.9\\[0.5ex]

Li et al., Alexnet \cite{li16} & 49.7 &33.6 &30.8 &19.9 &13 &40.5 &54.3 &37.4 &14.8 &39.8 &9.4 &28.8 &38.1 &49.8 &14.5 &24.0 &27.1 &12.1 &42.3 &39.7 &31.0\\[0.5ex]

Li et al., VGG16 \cite{li16} & 54.5 &47.4 &41.3 &20.8 &17.7 &51.9 &63.5 &46.1 &21.8 &57.1 &22.1 &34.4 &50.5 &61.8 &16.2 &29.9 &40.7 &15.9 &55.3 &40.2 &39.5\\[0.5ex]

WSDDN \cite{bilen16} & 46.4 &58.3 &35.5 &25.9 &14.0 &66.7 &53.0 &39.2 &8.9 &41.8 &26.6 &38.6 &44.7 &59.0 &10.8 &17.3 &40.7 &49.6 &56.9 &50.8 &39.3\\[0.5ex]
 
\hline 
WCCN\_2stage\_Alexnet & 43.5	&56.8	&34.1	&19.2	&13.4	&63.1	&51.5&	33.1	&5.8	&39.3	&19.6	&32.9&	46.2	&56.1&	11.2	&17.5	&38.5	&45.7&	52.6&	43.3 &36.2\\[0.5ex]

WCCN\_2stage\_VGG16 & 48.2	&58.9	&37.3	&27.8	&15.3	&69.8	&55.2&	41.1	&10.1	&42.7&	28.6	&40.4	&47.3&	62.3&	12.9	&21.2	&44.3	&52.2	&59.1&	53.1		&41.4\\[0.5ex]

WCCN\_3stage\_Alexnet & 43.9	&57.6	&34.9&	21.3	&14.7&	64.7	&52.8	&34.2	&6.5&	41.2	&20.5	&33.8&	47.6	&56.8	&12.7	&18.8	&39.6	&46.9&	52.9	&45.1		&37.3\\[0.5ex]

\textbf{WCCN\_3stage\_VGG16} & 49.5	&60.6	&38.6	&29.2&	16.2	&70.8	&56.9	&42.5	&10.9	&44.1	&29.9	&42.2	&47.9	&64.1	&13.8	&23.5	&45.9	&54.1	&60.8	&54.5	&	\textbf{42.8}\\[1ex]
\hline
\end{tabular}
}
\vspace{2pt}
 \caption{Detection average precision (\%) on the \textbf{PASCAL VOC 2007} dataset test set.}
  \label{tab:1}
\end{table*}

\begin{table*}[!htbp]
  \centering
  \resizebox{\textwidth}{!}{
  \begin{tabular}{  |c | c c c c c c c c c c c c c c c c c c c c| c| } 
 \hline
 Method & aero & bike & bird & boat & bottle & bus & car & cat & chair & cow & table & dog & horse & mbike & person & plant & sheep & sofa & train & tv & mAP\\ [0.5ex]
 \hline

WSDDN \cite{bilen16} & 95.0 & 92.6 & 91.2 & 90.4 & 79.0 & 89.2 & 92.8 & 92.4 &78.5 &90.5 &80.4 &95.1 &91.6 &92.5 &94.7 &82.2 &89.9 &80.3 &93.1 &89.1 &89.0\\[0.5ex]

Oquab et al. \cite{Oquab14} & 88.5 &81.5 &87.9 &82.0 &47.5& 75.5& 90.1 &87.2& 61.6& 75.7& 67.3& 85.5& 83.5 &80.0 &95.6 &60.8& 76.8 &58.0 &90.4 &77.9 &77.7\\[0.5ex]

SPPnet \cite{SPPNET} & $-$ &$-$ & $-$& $-$& $-$ &$-$ &$-$ &$-$& $-$ &$-$ &$-$& $-$ &$-$& $-$ &$-$& $-$& $-$& $-$& $-$& $-$ &82.4\\[0.5ex]

Alexnet \cite{bilen16} & 95.3 &90.4 &92.5 &89.6 &54.4 &81.9 &91.5 &91.9 &64.1 &76.3 &74.9 &89.7 &92.2 &86.9 &95.2 &60.7 &82.9 &68.0 &95.5 &74.4 &82.4\\[0.5ex]

VGG16-net \cite{vgg} & $-$ &$-$ & $-$& $-$& $-$ &$-$ &$-$ &$-$& $-$ &$-$ &$-$& $-$ &$-$& $-$ &$-$& $-$& $-$& $-$& $-$& $-$ &89.3\\[0.5ex]
\hline

WCCN\_2stage\_Alexnet & 92.8	&90.3	&89.3	&88.2	&80.4&	89.4&	90	&90.4&	75.3&	88.1&	80.1&	91.3&	89.1&	88.3	&91.2	&80.6	&88.5	&77.8	&92.2	&88.7	&	87.1\\[0.5ex]

WCCN\_2stage\_VGG16 & 93.4	&93.7&	92&	91	&83.1	&91.5&	92.7&	93.5&	79.3	&90.7	&83.1	&96.9&	92.9	&91.2	&95.9	&82.4	&90.3&	81.3&	95.1	&88.3	&	89.9\\[0.5ex]

WCCN\_3stage\_Alexnet & 93.1	&91.1	&89.6	&88.9	&81	&89.6	&90.7	&91.2&	76.4	&89.2	&80.8	&92.2	&90.1&	89	&92.7	&82	&89.3&	78.1	&92.8&89.1	&	87.8\\[0.5ex]

\textbf{WCCN\_3stage\_VGG16} & 94.2	&94.8&	92.8&	91.7&	84.1&	93	&93.5&	93.9&	80.7&	91.9&	85.3&	97.5&	93.4	&92.6&	96.1&	84.2&	91.1&	83.3&	95.5&	89.6	&	\textbf{90.9}\\[1ex]
\hline
\end{tabular}
}
\vspace{2pt}
 \caption{Classification average precision (\%) on the \textbf{PASCAL VOC 2007} test set.}
  \label{tab:2}
\end{table*}

\begin{table*}[!htbp]
  \centering
  \resizebox{\textwidth}{!}{
  \begin{tabular}{  |c | c c c c c c c c c c c c c c c c c c c c| c| } 
 \hline
 Method & aero & bike & bird & boat & bottle & bus & car & cat & chair & cow & table & dog & horse & mbike & person & plant & sheep & sofa & train & tv & mAP\\ 
 \hline

Bilen et al. \cite{bilen15} & 66.4 &59.3 &42.7 &20.4 &21.3& 63.4& 74.3& 59.6& 21.1 &58.2 &14.0 &38.5 &49.5 &60.0 &19.8& 39.2 &41.7 &30.1 &50.2 &44.1 &43.7\\ [0.5ex]

Cinbis et al. \cite{cinbis} & 65.3 &55.0& 52.4& 48.3 &18.2 &66.4 &77.8 &35.6& 26.5 &67.0 &46.9 &48.4 &70.5 &69.1& 35.2 &35.2 &69.6 &43.4 &64.6 &43.7 &52.0\\[0.5ex]

Wang et al. \cite{wang14} &80.1& 63.9& 51.5& 14.9& 21.0& 55.7& 74.2 &43.5 &26.2 &53.4 &16.3 &56.7 &58.3& 69.5 &14.1 &38.3 &58.8 &47.2 &49.1 &60.9 &48.5\\[0.5ex]

Li et al., Alexnet \cite{li16} & 77.3 &62.6& 53.3& 41.4 &28.7 &58.6& 76.2 &61.1 &24.5 &59.6 &18.0 &49.9 &56.8& 71.4 &20.9 &44.5& 59.4& 22.3& 60.9& 48.8 &49.8\\[0.5ex]

Li et al., VGG16 \cite{li16} & 78.2 &67.1& 61.8 &38.1& 36.1& 61.8 &78.8 &55.2& 28.5 &68.8 &18.5& 49.2 &64.1 &73.5& 21.4 &47.4& 64.6 &22.3 &60.9 &52.3 &52.4\\[0.5ex]

WSDDN \cite{bilen16} & 65.1 &63.4& 59.7 &45.9 &38.5 &69.4 &77.0& 50.7 &30.1 &68.8 &34.0 &37.3& 61.0 &82.9& 25.1 &42.9& 79.2 &59.4& 68.2 &64.1 &56.1\\[0.5ex]
 
\hline

WCCN\_2stage\_Alexnet & 78.4&	66.4&	58.2&	38.1&	34.9	&60.1	&77.8&	53.8&	26.6&	66.5	&18.7	&47.3&	62.8&	73.5	&20.4&	45.2	&64	&21.6	&59.9	&51.6	&	51.3\\[0.5ex]

WCCN\_2stage\_VGG16 & 81.2	&70	&62.5&	41.7&	38.2&	63.4&	81.1	&57.7	&30.4&	70.3&	21.7&	51&	65.9&	75.7&	23.9&	47.9&	67.5	&25.6	&62.4	&53.9	&	54.6\\[0.5ex]

WCCN\_3stage\_Alexnet & 79.7	&68.1&	60.4&	38.9&	36.8&	61.1&	78.6&	56.7&	27.8&	67.7	&20.3	&48.1&	63.9	&75.1	&21.5	&46.9	&64.8	&23.4	&60.2&	52.4	&	52.6\\[0.5ex]

\textbf{WCCN\_3stage\_VGG16} & 83.9&	72.8	&64.5&	44.1&	40.1	&65.7	&82.5	&58.9	&33.7	&72.5&	25.6	&53.7&	67.4	&77.4	&26.8	&49.1	&68.1	&27.9	&64.5	&55.7&		\textbf{56.7}\\[1ex]
\hline
\end{tabular}
}
\vspace{2pt}
 \caption{Correct localization (\%) on \textbf{PASCAL VOC 2007} on positive (CorLoc) trainval set.}
  \label{tab:3}
\end{table*}

\subsection{Datasets and metrics}
The experiments for our proposed methods are extensively done on the PASCAL VOC 2007, 2010, 2012 datasets and also ILSVRC 2013, 2014 which are large scale datasets for objects. The PASCAL VOC is more common dataset to evaluate weakly supervised object detection approaches. The VOC datasets have 20 categories of objects, while ILSVRC dataset has 200 categories which we targeted also for weakly supervised object classification and localization. In all of the mentioned datasets, we incorporate the standard train, validation and test set.

\textbf{Experimental metrics:} To measure the object detection performance, average precision (AP) and correct localization (CorLoc) is used. Average precision is the standard metric from PASCAL VOC which takes a bounding box as a true detection where it has intersection-over-union (IoU) of more than 50$\%$ with ground-truth box. The \textit{Corloc} is the fraction of positive images that the method obtained correct location for at least one object instance in the image. For the object classification, also we use PASCAL VOC standard average precision.

\subsection{Experimental and implementation details }
We have evaluated both of our proposed cascaded CNN with two architectures: Alexnet \cite{alexnet} and VGG-16 \cite{vgg}. In each case, the network has been pre-trained on ImageNet dataset \cite{imagenet}. Since the multiple stages of cascades contain different CNN networks losses, in the following we explain details of each part separately to have better overview of the implementation.
\\
\textbf{CNN architectures:}

\textbf{1.~Loc Net}: Inspired by \cite{gap_paper}, we removed fully-connected layers from each of Alexnet or VGG-16 and replaced them by two convolutional layers and one global pooling layer. So for the Alexnet, the layers after \textit{conv5} layer have been removed and for VGG-16 after \textit{conv5-3}. For global pooling layer, we have tested average and max pooling methods and we found that global average pooling performs better than maximum pooling. For the training loss criteria of this part of network, we use a simple sum of \textit{C} (number of classes) binary logistic regression losses, similar to \cite{laptev15}.

\textbf{2.~Seg Net}: This part of network is second stage in the 3-stage cascaded network and is well-known fully convolutional network for segmentation task. The convolutional part is shared with the other stages which comes from the first stage and additional fully-connected layers and a deconvolutional layer is used to produce segmentation map. The loss function is explained in section 3. Since this loss is provided by weak supervision, part of the supervision is obtained from the last stage in form of best initial regions of object instances.

\textbf{3.~MIL Net}: This last stage uses the shared convolutional feature maps as initial layers to train two fully-connected layers with size of 4096 and a label prediction layer. Using the the selected candidate bounding boxes from last stage (first stage in the 2stages cascade case and second stage in the 3stages cascade), it trains the multiple instance learning loss to select the best sample for each object presented in an image.

\textbf{Implementation details:} We use MatConvNet \cite{matconv} as CNN toolkit and all the networks are trained on a Geforce Titan X GPU. During the training time, images have been re-sized to multiple scale of images ($\{480, 576, 688, 84, 1200\}$) with respect to the original aspect ratio. The learning rate for the CNN networks is 0.0001 for 20 epochs and batch size of 100. For each image, we use 2000 object proposals generated by EdgeBox or SelectiveSearch algorithms. At the last stage, we select 10 boxes for each object instance in each iteration for training multiple instance learning. To use Fast-RCNN detection with the ground-truths that are obtained by our methods, we set the number of iterations to 40K. For selecting the candidate boxes in our pipelines, we use a thresholding method like \cite{gap_paper} for weakly localization.

\begin{figure*}[htbp]
 \centering
 \includegraphics[width=500pt]{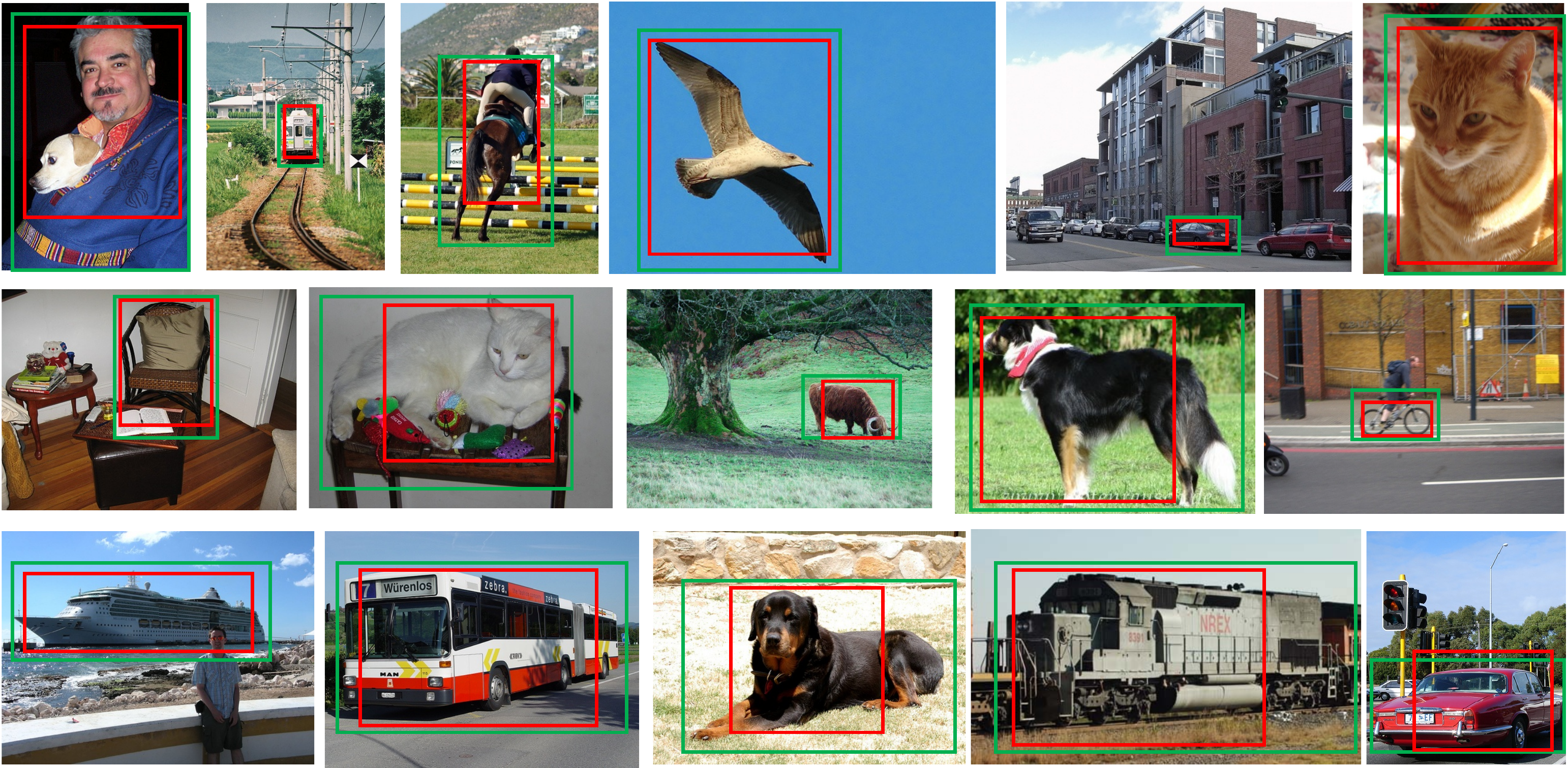}
 \caption{Examples of our object detection results. Green bounding boxes are ground-truth annotations and red boxes are positive detection. Images are sampled from PASCAL VOC 2007 test set.}
  \label{fig:2}
 \end{figure*}

\subsection{Detection performance}
\textbf{Comparison with the state-of-the-art:} We evaluate the detection performance of our method in this section. To compare our approach, methods which use deep learning pipelines \cite{bilen16, li16} or multiple instance learning algorithms \cite{cinbis} or clustering based approaches \cite{bilen15} are studied.

Tables \ref{tab:1}, \ref{tab:4}, \ref{tab:5} present results on PASCAL VOC 2007, 2010, 2012 for object detection on test sets with average precision measurement. It can be observed that by using the weakly supervision setup, we achieved the best performance among of all other recent methods. Our approaches does not incorporate any sophisticated clustering or optimized initialization step, and all the steps are trained together via an end-to-end learning of deep neural networks. There is a semantic relationship between improvements gain using different CNN architectures in our networks in comparison with using the same CNNs in other methods. We have almost the same improvement with two different architectures over other methods.

The localization performance with {CorLoc} metric is also shown in Table \ref{tab:3} on PASCAL VOC 2007. Our best performance is 56.7\% which is achieved by 3stage cascade network using VGG-16 architecture. However, our network with the Alexnet outperformed the other methods using similar network architectures with same number of layers and other non deep learning methods. Most of the other works use CNNs as some part of their pipeline, not in an end-to-end scheme or use it simply as a feature extractor. Differently, our cascaded deep networks will bring multiple concepts together in a single training method, learn better appearance model and feature representation for objects under weakly supervision circumstances.

\begin{table}[htb] 
\begin{center}
\resizebox{8.4cm}{!} {
\begin{tabular}{|l|c|c|c|}
\hline 
Method &  VOC2010 & VOC2012 & ILSVRC 2013  \\

\hline\hline
Cinbis et al. \cite{cinbis} & 27.4& $-$ & $-$ \\
Wang et al. \cite{wang14}& $-$& $-$  & 6.0 \\
Li et al., Alexnet \cite{li16}& 21.4& 22.4 & 7.7  \\
Li et al., VGG16 \cite{li16} & 30.7& 29.1 & 10.8 \\
WSDDN \cite{bilen16} & 36.2& $-$  & $-$ \\
\hline
WCCN\_2stage\_Alexnet& 27.6 & 27.3 & 9.1 \\
{WCCN\_2stage\_VGG16}& {37.8}& {36.4}  & {14.6}\\
{WCCN\_3stage\_Alexnet}& {28.8}& {28.4}  & {9.8}\\
\textbf{WCCN\_3stage\_VGG16}& \textbf{39.5}& \textbf{37.9}  & \textbf{16.3}\\
\hline
\end{tabular}}
\end{center}
\caption{Detection performance (\%) comparison on VOC 2010, 2012 test set and ILSVRC 2013 validation set.}
  \label{tab:4}
\end{table}

We also compared our object detector results on ILSVRC'13 only with \cite{li16,wang14}, since no other weakly supervised object detector methods have been tried on this dataset. Results are shown in Table \ref{tab:4} and similar to previous tests, we achieved better number in performance. Since, some part of our work is inspired by GAP networks from \cite{gap_paper}, we compared our weakly supervised localization on the ILSVRC'14 dataset following their experimental setups and the results are in Table \ref{tab:6}.

\begin{figure*}[htb]
\centering
\resizebox{15cm}{!}{
\begin{tikzpicture}[scale=0.75]
  \centering
  \begin{axis}[ybar, axis on top,
        height=4cm, width=16cm,
        bar width=0.15cm,
        tick align=inside,
        major grid style={draw=white},
        enlarge y limits={value=.1,upper},
        ymin=0.1, ymax=0.72,
        axis x line*=bottom,
        axis y line*=left,
        y axis line style={opacity=0},
        tickwidth=0pt,
        enlarge x limits=true,
        legend style={
            at={(0.5,-0.35)},
            anchor=north,
            legend columns=-1,
            /tikz/every even column/.append style={column sep=0.25cm}
        },
        ylabel={mAP},
   symbolic x coords={aero,bike,bird,boat,bottle,bus,car,cat,chair,cow,table,dog,horse,mbike,person,plant,sheep,sofa,train,tv},
    xtick=data,
    xticklabel style={
        inner sep=0pt,
        anchor=north east,
        rotate=45
    },
    ]
    \addplot coordinates {(aero,0.495) (bike,0.606) (bird,0.386) (boat,0.292) (bottle,0.162) (bus,0.708) (car,0.569)  (cat,0.425) (chair,0.109) (cow,0.441) (table,0.299) (dog,0.422) (horse,0.479) (mbike,0.641) (person,0.138) (plant,0.235) (sheep,0.459) (sofa,0.541) (train,0.608) (tv,0.545) }; 
    
    \addplot coordinates {(aero,0.501) (bike,0.609) (bird,0.379) (boat,0.301) (bottle,0.167) (bus,0.708) (car,0.575)  (cat,0.428) (chair,0.114) (cow,0.442) (table,0.303) (dog,0.423) (horse,0.481) (mbike,0.64) (person,0.149) (plant,0.239) (sheep,0.465) (sofa,0.547) (train,0.611) (tv,0.548) };
    \legend{WCCN,Fast-RCNN (with our pseudo GT)}
  \end{axis} 
\end{tikzpicture}}
\caption{Comparison between our detection full pipeline and training Fast\_RCNN using pseudo ground-truth bounding boxes extracted by our method.} \label{fig:detect}
\end{figure*}
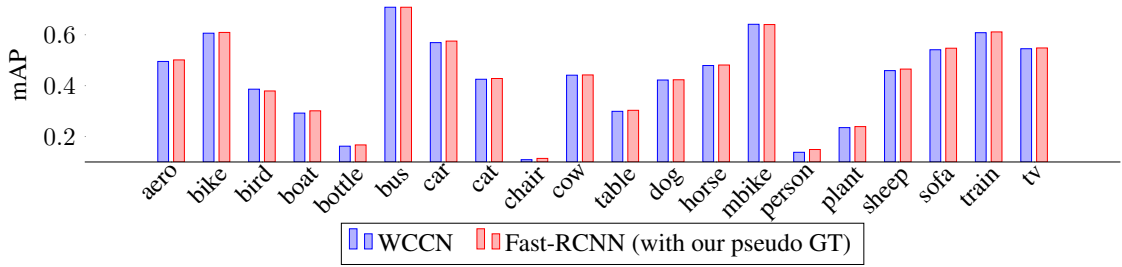

\textbf{Object detection training:} We compared our full detection pipeline with the state-of-the-art detection method, Fast\_RCNN implemented in Caffe \cite{caffe}. Since the Fast\_RCNN \cite{fastRCNN} is a supervised method, we use the pseudo ground-truth (GT) bounding boxes which are generated by our cascaded networks. By our experiments, In the Fig.\ref{fig:detect}, it is shown that the Fast\_RCNN pipeline can also perform good results with our input bounding boxes. Fast\_RCNN trained by our generated GT performs slightly better than our detection full pipeline on the average precision of PASCAL VOC 2007 test set (0.3\%). The main goal of this work is to find the most representative and discriminative samples that signify the existing categories in each image.

\begin{table}
\begin{center}
\resizebox{6cm}{!} {
\begin{tabular}{|l|c|}
\hline
Method &  Top-1 error \\
\hline\hline
Alexnet & 65.17 \\
VGG16 & 61.12  \\
Alexnet-GAP \cite{gap_paper}& 63.75  \\
VGG16-GAP \cite{gap_paper} & 57.20 \\
\hline
WCCN\_2stage\_Alexnet& 62.2 \\
\textbf{WCCN\_2stage\_VGG16}& \textbf{55.6} \\
\hline
\end{tabular}}
\end{center}
\caption{Detection top-1 error (\%) on ILSVRC'14 validation set}
  \label{tab:5}
\end{table}

\begin{table}
\begin{center}
\resizebox{6cm}{!} {
\begin{tabular}{|l|c|}
\hline
Method &  Top-1 error \\
\hline\hline
Alexnet & 42.6 \\
VGG16 & 31.2  \\
Alexnet-GAP \cite{gap_paper}& 44.9  \\
VGG16-GAP \cite{gap_paper} & 33.4 \\
\hline
WCCN\_2stage\_Alexnet& 41.2 \\
\textbf{WCCN\_2stage\_VGG16}& \textbf{30.4} \\
\hline
\end{tabular}}
\end{center}
\caption{Classification top-1 error (\%) on ILSVRC'14 validation set}
  \label{tab:6}
\end{table}

\textbf{Object proposals:} In our work, we evaluated the effect of different unsupervised object proposals generator. EdgeBox \cite{edgebox} and SelectiveSearch \cite{SS} are compared based on the detector trained by our networks. According to the results on the VOC 2007 detection test set, by training 2stage cascade using Alexnet with Edgebox, approximately 1.5\% improvement can be obtained over SelectiveSearch. Similar to the other works like \cite{bilen16,SPPNET}, EdgeBox performs better with CNN based object detectors.

\subsection{Classification performance}
 Our proposed network design has dual purposes: object detection and classification in a weakly supervision manner. Obviously the structure of our cascade is helpful for training classification pipeline on images with multiple objects and minimum supervision of labels. We evaluated our method on PASCAL VOC 2007 and ILSVRC 2014. The performance is compared with other approaches which use novel methods in deep learning for classification on these datasets. 
 
Table \ref{tab:2} presents the comparison on VOC 2007 with different CNN architectures for all of the methods. Since first stage of our cascade is similar to \cite{gap_paper}, we show the result of classification on ILSVRC'14, the large scale dataset for classification, in Table \ref{tab:6}.

\subsection{Cascade Architecture  Study}
If an ablation study would be interesting over the stages of proposed cascades, it can be noticed that all of the results show how each of the proposed cascades can affect the performance in detection or classification. Each stage in our multi-stage cascaded CNN can be analyzed by comparison with the CNN-based methods in same context. Training the stage with multiple instance loss can improve learning the best sample of each category over other works \cite{gap_paper, bilen16}.
It can be observed that adding the stage of segmentation to exploit better regions can outperform the two-stage cascade. Adding segmentation stage has impact on finding more accurate initial guess of object locations. For an instance of using the segmentation stage by Alexnet architecture, cascaded network improves almost 2.5\% on detection and 2\% on classification in PASCAL VOC 2007.


\section{Conclusion}
Our idea of weak cascaded convolutional networks (WCCN) is about the approaches of cascaded CNNs for weakly supervised visual learning tasks like object detection, localization and classification. In this work, we proposed two multi-stage cascaded networks with different loss functions in each stage to conclude a better pipeline of deep convolutional neural network learning with weak supervision of object labels on images. Our insight was a paradigm of multi-task learning effectiveness using deep neural networks. We proved that our multi-task learning approaches that incorporate localization, multiple instance learning and weakly supervised segmentation of object regions achieve the state-of-the-art performance in weakly supervised object detection and classification. The extensive experiments for object detection and classification tasks on various datasets like PASCAL VOC 2007, 2010, 2012 and also large scale datasets, ILSVRC 2013, 2014 present the full capability of the proposed method.

\subsection*{Acknowledgements}
This work was supported by DBOF PhD scholarship, KU Leuven CAMETRON project. The authors would like to thank Nvidia for GPU donation.

{\small
\bibliographystyle{ieee}
\bibliography{egbib}
}

\end{document}